\documentclass[11pt,a4paper]{article}
\usepackage[hyperref]{acl2020}
\usepackage{times}
\usepackage{latexsym}

\usepackage{graphicx}
\usepackage{url}
\usepackage{multirow}
\usepackage{booktabs}

\usepackage{microtype}

\aclfinalcopy % Uncomment this line for the final submission
\title{Towards Conversational Recommendation over Multi-Type Dialogs}
\author{
	Zeming Liu\textsuperscript{1}\thanks{\quad This work was done at Baidu.}, Haifeng Wang\textsuperscript{2}, Zheng-Yu Niu\textsuperscript{2}, Hua Wu\textsuperscript{2}, Wanxiang Che\textsuperscript{1}\thanks{\quad Corresponding author: Wanxiang Che.}, Ting Liu\textsuperscript{1}\\
	\textsuperscript{1}Research Center for Social Computing and Information Retrieval,\\
	Harbin Institute of Technology, Harbin, China \\
	\textsuperscript{2}Baidu Inc., Beijing, China \\
	{\tt \{zmliu, car, tliu\}@ir.hit.edu.cn} \\
	{\tt \{wanghaifeng, niuzhengyu, wu\_hua\}@baidu.com} \\
}
\date{}
\begin{document}
	\maketitle
	\begin{abstract}
		We propose a new task of conversational recommendation over multi-type dialogs, where the bots can proactively and naturally lead a conversation from a non-recommendation dialog (e.g., QA) to a recommendation dialog, taking into account user's interests and feedback. To facilitate the study of this task, we create a human-to-human Chinese dialog dataset \emph{DuRecDial} (about 10k dialogs, 156k utterances), which contains multiple sequential dialogs for every pair of a recommendation seeker (user) and a recommender (bot). In each dialog, the recommender proactively leads a multi-type dialog to approach recommendation targets and then makes multiple recommendations with rich interaction behavior. This dataset allows us to systematically investigate different parts of the overall problem, e.g., how to naturally lead a dialog, how to interact with users for recommendation. Finally we establish baseline results on \emph{DuRecDial} for future studies.\footnote{Dataset and codes are publicly available at https://github.com/PaddlePaddle/models/\\tree/develop/PaddleNLP/Research/ACL2020-DuRecDial.}
	\end{abstract}
	
	\section{Introduction}
	%1. conversational rec 介绍， 前人工作总结
	%2. 问题
	%3. 新任务设定：主动推荐(用户没有明确预期，从某个场景 跳到推荐)，数据集的设计思路:冷启动，profile 更新， 对话逻辑交互性（回绝，问答，主动切换，接受），对话理由（话术）， session中多个goal， 多领域，多种对话类型, which consists of four key components, goal planning, recommendation target selection, goal-driven conversation generation, and persona update
	%4. 我们还提供了baselines
	%IJCAI  Learning Conversational Systems that Interleave Task and Non-Task Content
	%Third, the dialogue initiative is rigidly prescribed. Es- sentially, the user makes a request, the system asks for the missing information, the user supplies that information, the system (eventually) confirms the action to be performed, the user agrees or disconfirms, etc.
	
	In recent years, there has been a significant increase in the work of conversational recommendation due to the rise of voice-based bots \cite{Christakopoulou2016,Li2018,Reschke2013,Warnestal2005}. They focus on how to provide high-quality recommendations through dialog-based interactions with users. These work fall into two categories: (1) task-oriented dialog-modeling approaches \cite{Christakopoulou2016,Sun2018,Warnestal2005}; (2) non-task dialog-modeling approaches with more free-form interactions \cite{Kang2019,Li2018}. Almost all these work focus on a single type of dialogs, either task oriented dialogs for recommendation, or recommendation oriented open-domain conversation. Moreover, they assume that both sides in the dialog (especially the user) are aware of the conversational goal from the beginning.
	
	\begin{figure*}[t]
		\centering\includegraphics[ height=3.8in]{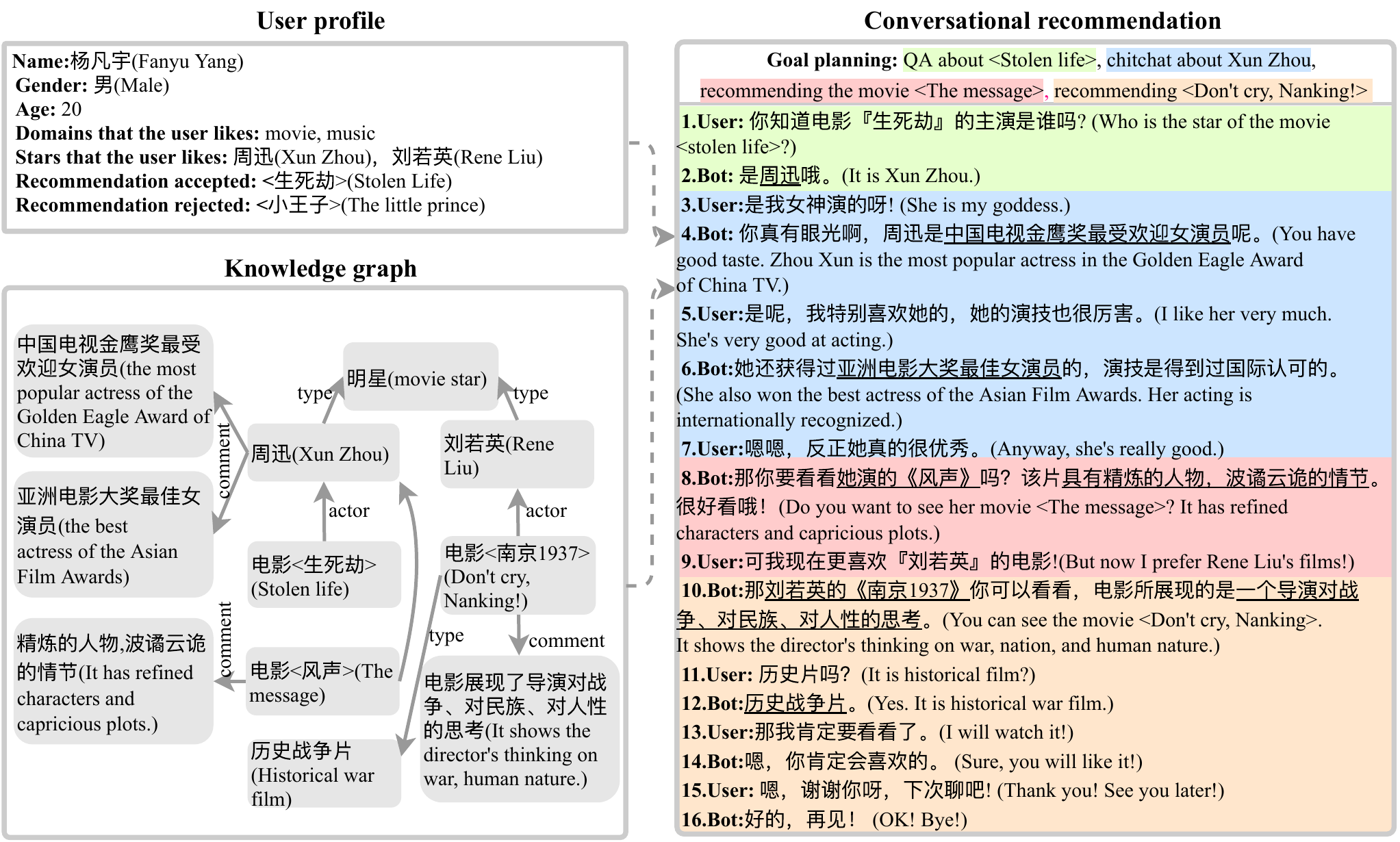}
		\caption{A sample of conversational recommendation over multi-type dialogs. The whole dialog is grounded on knowledge graph and a goal sequence, while the goal sequence is planned by the bot with consideration of user's interests and topic transition naturalness. Each goal specifies a dialog type and a dialog topic (an entity). We use different colors to indicate different goals and use underline to indicate knowledge texts. }\label{fig:example}%, which consists of a QA sub-dialog, a recommendation sub-dialog and a task sub-dialog. 
	\end{figure*}
	
	In many real-world applications, there are multiple dialog types in human-bot conversations (called \emph{multi-type dialogs}), such as chit-chat, task oriented dialogs, recommendation dialogs, and even question answering \cite{Alex2018,Wang2014,XiaoIce2018}. Therefore it is crucial to study how to proactively and naturally make conversational recommendation by the bots in the context of multi-type human-bot communication. For example, the bots could proactively make recommendations after question answering or a task dialog to improve user experience, or it could lead a dialog from chitchat to approach a given product as commercial advertisement. However, to our knowledge, there is less previous work on this problem. 
	
	To address this challenge, we present a novel task, conversational recommendation over multi-type dialogs, where we want the bot to proactively and naturally lead a conversation from a non-recommendation dialog to a recommendation dialog. For example, in Figure \ref{fig:example}, given a starting dialog such as question answering, the bot can take into account user's interests to determine a recommendation target (the movie $<$The message$>$) as a long-term goal, and then drives the conversation in a natural way by following short-term goals, and completes each goal in the end. Here each goal specifies a dialog type and a dialog topic. Our task setting is different from previous work \cite{Christakopoulou2016,Li2018}. First, the overall dialog in our task contains multiple dialog types, instead of a single dialog type as done in previous work. Second, we emphasize the initiative of the recommender, i.e. the bot proactively plans a goal sequence to lead the dialog, and the goals are unknown to the users. When we address this task, we will encounter two difficulties: (1) how to proactively and naturally lead a conversation to approach the recommendation target, (2) how to iterate upon initial recommendation with the user. 
	
	To facilitate the study of this task, we create a human-to-human \textbf{rec}ommendation oriented multi-type Chinese \textbf{dial}og dataset at Bai\textbf{du} (\textbf{DuRecDial}). In \emph{DuRecDial}, every dialog contains multi-type dialogs with natural topic transitions, which corresponds to the first difficulty. Moreover, there are rich interaction variability for recommendation, corresponding to the second difficulty. In addition, each seeker has an explicit profile for the modeling of personalized recommendation, and multiple dialogs with the recommender to mimic real-world application scenarios. %, which supports the study of user preference modeling. 
	
	To address this task, inspired by the work of \citet{Xu2020}, we present a \textbf{m}ulti-\textbf{g}oal driven \textbf{c}onversation \textbf{g}eneration framework (\textbf{MGCG}) to handle multi-type dialogs simultaneously, such as QA/chitchat/recommendation/task etc.. It consists of a goal planning module and a goal-guided responding module. The goal-planning module can conduct dialog management to control the dialog flow, which determines a recommendation target as the final goal with consideration of user's interests and online feedback, and plans appropriate short-term goals for natural topic transitions. To our knowledge, this goal-driven dialog policy mechanism for multi-type dialog modeling is not studied in previous work. The responding module produces responses for completion of each goal, e.g., chatting about a topic or making a recommendation to the user. We conduct an empirical study of this framework on \emph{DuRecDial}.
	
	This work makes the following contributions: 
	\begin{itemize}
		\item We identify the task of conversational recommendation over multi-type dialogs. 
		\item To facilitate the study of this task, we create a novel dialog dataset \emph{DuRecDial}, with rich variability of dialog types and domains as shown in Table \ref{table:datasets}.
		\item We propose a conversation generation framework with a novel mixed-goal driven dialog policy mechanism.
	\end{itemize}
	
	\begin{table*}
		\centering
		\small
		\begin{tabular}{ p{4.4cm} r r p{1.8cm} p{3.4cm} r  } 
			\toprule[1.0pt]
			Datasets$\downarrow$ Metrics$\rightarrow$  & \#Dialogs & \#Utterances & Dialog types  &  Domains  & User profile \\ 
			\toprule[1.0pt]
			Facebook\_Rec \cite{Dodge2016} & 1M & 6M & Rec. & Movie&  No \\  
			REDIAL \cite{Li2018} & 10k& 163k& Rec., chitchat& Movie&  No \\ 
			%EMNLP \cite{Zhou2018} & 1& 1& Rec., QA& Movie& 1  \\ \hline
			GoRecDial \cite{Kang2019} & 9k& 170k& Rec.& Movie&   Yes \\ 
			OpenDialKG \cite{Moon2019}& 12k& 143k& Rec.& Movie, book&   No \\ 
			\hline
			CMU DoG \cite{Zhou2018}& 4k& 130k& Chitchat & Movie&   No  \\ 
			IIT DoG \cite{Moghe2018}& 9k& 90k& Chitchat & Movie&   No  \\ 
			Wizard-of-wiki \cite{Dinan2019}& 22k& 202k& Chitchat & 1365 topics from Wikipedia &   No  \\ 
			OpenDialKG \cite{Moon2019}& 3k& 38k& Chitchat & Sports, music&   No \\ 
			DuConv \cite{Wu2019} & 29k& 270k& Chitchat & Movie&   No  \\ 
			KdConv \cite{Zhou2020}& 4.5k& 86k& Chitchat & Movie, music, travel&   No  \\ 
			%MultiWOZ \cite{Budzianowski2018} & 10,438& 115,434 & Task& Restaurant, hotel, attraction, taxi, train, hospital, police&   No  \\ \hline
			\hline
			DuRecDial & 10.2k& 156k & Rec., chitchat, QA, task & Movie, music, movie star, food, restaurant, news, weather&   Yes\\ \bottomrule[1.0pt]
		\end{tabular}
		\caption{Comparison of our dataset \emph{DuRecDial} to recommendation dialog datasets and knowledge grounded dialog datasets. ``Rec.'' stands for recommendation.}
		\label{table:datasets}
	\end{table*}
	
	\section{Related Work}
	\textbf{Datasets for Conversational Recommendation} To facilitate the study of conversational recommendation, multiple datasets have been created in previous work, as shown in Table \ref{table:datasets}. The first recommendation dialog dataset is released by \citet{Dodge2016}, which is a synthetic dialog dataset built with the use of the classic MovieLens ratings dataset and natural language templates. \citet{Li2018} creates a human-to-human multi-turn recommendation dialog dataset, which combines the elements of social chitchat and recommendation dialogs. \citet{Kang2019} provides a recommendation dialogue dataset with clear goals, and \citet{Moon2019} collects a parallel Dialog$\leftrightarrow$KG corpus for recommendation. Compared with them, our dataset contains multiple dialog types, multi-domain use cases, and rich interaction variability. 
	
	\textbf{Datasets for Knowledge Grounded Conversation} As shown in Table \ref{table:datasets}, CMU DoG \cite{Zhou2018} explores two scenarios for Wikipedia-article grounded dialogs: only one participant has access to the document, or both have. IIT DoG \cite{Moghe2018} is another dialog dataset for movie chats, wherein only one participant has access to background knowledge, such as IMDB's facts/plots, or Reddit's comments. \citet{Dinan2019} creates a multi-domain multi-turn conversations grounded on Wikipedia articles. OpenDialKG \cite{Moon2019} provides a chit-chat dataset between two agents, aimed at the modeling of dialog logic by walking over knowledge graph-Freebase. \citet{Wu2019} provides a Chinese dialog dataset-DuConv, where one participant can proactively lead the conversation with an explicit goal. KdConv \cite{Zhou2020} is a Chinese dialog dataset, where each dialog contains in-depth discussions on multiple topics. In comparison with them, our dataset contains multiple dialog types, clear goals to achieve during each conversation, and user profiles for personalized conversation. 

	\textbf{Models for Conversational Recommendation} Previous work on conversational recommender systems fall into two categories: (1) task-oriented dialog-modeling approaches in which the systems ask questions about user preference over predefined slots to select items for recommendation \cite{Christakopoulou2018,Christakopoulou2016,Lee2018,Reschke2013,Sun2018,Warnestal2005,Zhang2018}; (2) non-task dialog-modeling approaches in which the models learn dialog strategies from the dataset without predefined task slots and then make recommendations without slot filling \cite{Chen2019,Kang2019,Li2018,Moon2019,Zhou2018}. Our work is more close to the second category, and differs from them in that we conduct multi-goal planning to make proactive conversational recommendation over multi-type dialogs. 
	
	\textbf{Goal Driven Open-domain Conversation Generation} Recently, imposing goals on open-domain conversation generation models having attracted lots of research interests \cite{Moon2019,Li2018,Tang2019,Wu2019} since it provides more controllability to conversation generation, and enables many practical applications, e.g., recommendation of engaging entities. However, these models can just produce a dialog towards a single goal, instead of a goal sequence as done in this work. We notice that the model by \citet{Xu2020} can conduct multi-goal planning for conversation generation. But their goals are limited to in-depth chitchat about related topics, while our goals are not limited to in-depth chitchat. 
	
	\begin{figure}[t]
		\centering\includegraphics[width=2.7in,height=2.2in]{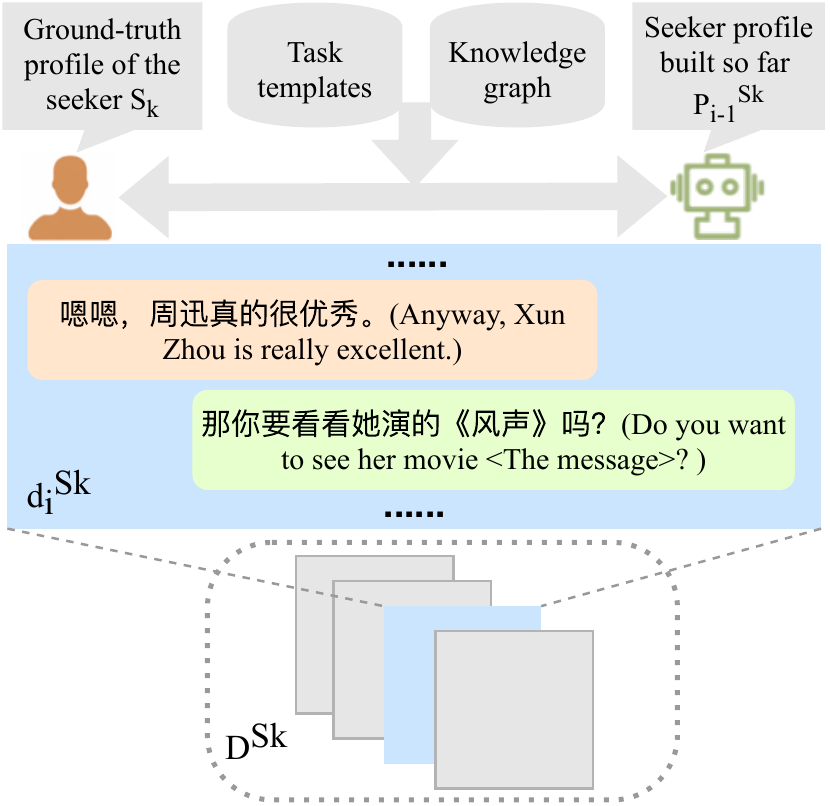}
		\caption{We collect multiple sequential dialogs $\{d^{s_k}_{i}\}$ for each seeker $s_k$. For annotation of every dialog, the recommender makes personalized recommendations according to task templates, knowledge graph and the seeker profile built so far. The seeker must accept/reject the recommendations.}\label{fig:taskdesign}
	\end{figure}
	
	\section{Dataset Collection\footnote{Please see Appendix 1. for more details.}}
	%In this section, we first present the motivation and design of our task. Then we describe how we collect the dataset \emph{DuRecDial} and provide detailed dataset statistics. 

	\subsection{Task Design}
	%seeker-recommnder互聊
	%seeker profile
	%goal 序列
	%主动引导
	%Here we formalize the task design of conversational recommendation over multi-type dialogs for the purposes of data collection. 
	We define one person in the dialog as the recommendation seeker (the role of users) and the other as the recommender (the role of bots). We ask the recommender to proactively lead the dialog and then make recommendations with consideration of the seeker's interests, instead of the seeker to ask for recommendation from the recommender. Figure \ref{fig:taskdesign} shows our task design. The data collection consists of three steps: (1) collection of seeker profiles and knowledge graph; (2) collection of task templates; (3) annotation of dialog data.\footnote{We have a strict data quality control process, please see Appendix 1.2 for more details.} Next we will provide details of each step.
	
	\textbf{Explicit seeker profiles} Each seeker is equipped with an explicit unique profile (a ground-truth profile), which contains the information of name, gender, age, residence city, occupation, and his/her preference on domains and entities. We automatically generate the ground-truth profile for each seeker, which is known to the seeker, and unknown to the recommender. We ask that the utterances of each seeker should be consistent with his/her profile. We expect that this setting could encourage the seeker to clearly and self-consistently explain what he/she likes/dislikes. In addition, the recommender can acquire seeker profile information only through dialogs with the seekers. % 对话结束后 会更新某些字段。
	
	\textbf{Knowledge graph} Inspired by the work of document grounded conversation \cite{Ghazvininejad2018,Moghe2018}, we provide a knowledge graph to support the annotation of more informative dialogs. We build them by crawling data from Baidu Wiki and Douban websites. Table \ref{table:statistics} presents the statistics of this knowledge graph.
	
	\textbf{Multi-type dialogs for multiple domains} We expect that the dialog between the two task-workers starts from a non-recommendation scenario, e.g., question answering or social chitchat, and the recommender should proactively and naturally guide the dialog to a recommendation target (an entity). The targets usually fall into the seeker's interests, e.g., the movies of the star that the seeker likes.
	
	Moreover, to be close to the setting in practical applications, we ask each seeker to conduct multiple sequential dialogs with the recommender. In the first dialog, the recommender asks questions about seeker profile. Then in each of the remaining dialogs, the recommender makes recommendations based on the seeker's preferences collected so far, and then the seeker profile is automatically updated at the end of each dialog. We ask that the change of seeker profile should be reflected in later dialogs. The difference between these dialogs lies in sub-dialog types and recommended entities. 
	
	\textbf{Rich variability of interaction} How to iterate upon initial recommendation plays a key role in the interaction procedure for recommendation. To provide better supervision for this capability, we expect that the task workers can introduce diverse interaction behaviors in dialogs to better mimic the decision-making process of the seeker. For example, the seeker may reject the initial recommendation, or mention a new topic, or ask a question about an entity, or simply accept the recommendation. The recommender is required to respond appropriately and follow the seeker's new topic.
	
	%[1]寒暄(Bot主动，根据给定的『聊天主题』寒暄，第一句问候要带User名字，聊天内容不要与『聊天时间』矛盾，聊天要自然，不要太生硬)-->[2]儿歌推荐(Bot主动，Bot使用给定的关于 『猫』 的某个信息当做推荐理由来推荐 『小猫小花猫，睡大觉。醒来喵喵喵。』，User回答的大概内容是『暂时不想聊儿歌』，鼓励具体回复的多样性)-->[3]关于动物的聊天(Bot主动，根据给定的动物信息聊 『猫』 相关内容，至少要聊2轮，避免话题切换太僵硬，不够自然)-->[4]动漫推荐(Bot主动，Bot使用动漫 『海绵宝宝特辑』 的部分『描述』当做推荐理由来主动推荐 『海绵宝宝特辑』, User拒绝，拒绝原因可以是『 看过、暂时不想看、对这个动漫不感兴趣 或 其他原因』；Bot使用动漫 『猫狗』 的『部分描述』当做推荐理由来主动推荐 『猫狗』，User先问动漫『主角、类型、描述、适合人群』中的一个或多个，Bot回答，最终User接受。注意：不要在一句话推荐两个动漫)-->[5]再见
	\begin{table}[t]
		\centering
		\small
		\begin{tabular}{ p{0.8in}  p{2.0in} } 
			\toprule[1.0pt]
			Goals & Goal description\\  
			\toprule[1.0pt]
			%问答 ( User 主动 ， 问 电影   『 生死劫 』   主演 是 谁 ? ， Bot 回答   『 周迅 』 ， User 满足 并 好评 ) --> [2
			%2] 关于 明星 的 聊天 ( Bot 主动 ， 根据 给定 的 明星 信息 聊   『 周迅 』   相关 内容 ， 至少 要 聊 2 轮 ， 避免 话题 切换 太 僵硬 ， 不够 自然 ) %[3] 电影 推荐 ( Bot 主动 ， Bot 使用   『 风声 』   的 某个 评论 当做 推荐 理由 来 推荐   『 风声 』 ， User 切换 话题 :   User 现在 更 喜欢   『 刘若英 』   的 电影 。 ) -
			%[4] 电影 推荐 ( Bot 主动 ， Bot 使用   『 南京1937 』   的 某个 评论 当做 推荐 理由 来 推荐   『 南京1937 』 ， User 先问 电影 『 国家 地区 、 导演 、 类型 、 主演 、 口碑 、 评分 』 中 的 一个 或 多个 ， Bot 回答 ， 最终 User 接受 ) 
			Goal1: QA (dialog type) about the movie $<$Stolen life$>$ (dialog topic)
			&The seeker takes the initiative, and asks for the information about the movie $<$Stolen life$>$; the recommender replies according to the given knowledge graph; finally the seeker provides feedback.
			\\\hline
			Goal2: chitchat about the movie star Xun Zhou
			&The recommender proactively changes the topic to movie star Xun Zhou as a short-term goal, and conducts an in-depth conversation; 
			\\\hline
			Goal3: Recommendation of the movie $<$The message$>$
			&The recommender proactively changes the topic from movie star to related movie $<$The message$>$, and recommend it with movie comments, and the seeker changes the topic to Rene Liu's movies; 
			\\\hline
			Goal4: Recommendation of the movie $<$Don't cry, Nanking!$>$
			&The recommender proactively recommends Rene Liu's movie $<$Don't cry, Nanking!$>$ with movie comments. The seeker tries to ask questions about this movie, and the recommender should reply with related knowledge. Finally the user accepts the recommended movie. 
			\\ 
			\bottomrule[1.0pt]
		\end{tabular}
		\caption{One of our task templates that is used to guide the workers to annotate the dialog in Figure \ref{fig:example}. We require that the recommendation target (the long-term goal) is consistent with the user's interests and the topics mentioned by the user, and short-term goals provide natural topic transitions to approach the long-term goal.}
		\label{table:templates}
	\end{table}
	
	%\begin{figure}[t]
	%		\centering\includegraphics[width=3.1in,height=3.4in]{Figure3.pdf}
	%		\caption{One of our task templates that is used to guide the workers to annotate the dialog in Figure \ref{fig:example}. Both goals and goal descriptions are prepared with the use of some rules and human annotation.}\label{fig:templates}
	%	\end{figure}
	
	\textbf{Task templates as annotation guidance} Due to the complexity of our task design, it is very hard to conduct data annotation with only high-level instructions mentioned above. Inspired by the work of MultiWOZ \cite{Budzianowski2018}, we provide a task template for each dialog to be annotated, which guides the workers to annotate in the way we expect them to be. As shown in Table \ref{table:templates}, each template contains the following information: (1) a goal sequence, where each goal consists of two elements, a dialog type and a dialog topic, corresponding to a sub-dialog. (2) a detailed description about each goal. We create these templates by (1) first automatically enumerating appropriate goal sequences that are consistent with the seeker's interests and have natural topic transitions and (2) then generating goal descriptions with the use of some rules and human annotation. 
	
	%{"situation": "聊天时间:2018-10-12 中午12:00，在学校 星期五", "conversation": ["[1]石家庄今天天气怎么样？", "石家庄2018年10月12日,多云，南风 最高气温20摄氏度，最低气温10摄氏度", "嗯嗯，怪不得有点冷呢", "[2]气温低建议你吃点辣的东西，重庆辣子鸡就很不错", "嗯嗯，它主要用了什么食材呢", "重庆辣子鸡，主要成分是鸡翅，配上花生、蒜苔、干辣椒等食材，吃完让人回味无穷", "嗯嗯，听起来很不错，有机会一定要去试试", "[3]说到辣子鸡，老酒川菜坊宽院子店（原体南店)这家店的重庆辣子鸡做的很不错", "那它家在哪儿呢？人均消费价格怎么样？", "老酒川菜坊宽院子店（原体南店）地址在裕华区体育南大街270号（省工商局招待所院内）,人均消费价格是49", "嗯嗯，评分怎么样？", "老酒川菜坊宽院子店（原体南店）评分3.9", "看来不错哦!明天可以去尝尝", "那需要帮您预定座位吗？", "好的", "那请您提供一下就餐人数和就餐时间", "明天中午，3个人", "好的，已为您预定成功", "好的，谢谢！我要去午休了，有空再聊", "[4]好的，再见"], "kg": [["石家庄", "2018-10-12", "多云, 南风, 最高气温:20℃, 最低气温:10℃"], ["重庆辣子鸡", "成分", "鸡翅、花生米、蒜苔、干辣椒、花椒、麻椒、盐、糖、料酒、蒜、姜。"], ["重庆辣子鸡", "类型", "热菜"], ["多云,", "南风, 最高气温:20℃, 最低气温:10℃ 适合吃", "重庆辣子鸡"], ["老酒川菜坊宽院子店（原体南店）", "特色菜", "重庆辣子鸡"], ["老酒川菜坊宽院子店（原体南店）", "评分", "3.9"], ["老酒川菜坊宽院子店（原体南店）", "人均价格", "49"], ["老酒川菜坊宽院子店（原体南店）", "地址", "裕华区体育南大街270号（省工商局招待所院内）"], ["老酒川菜坊宽院子店（原体南店）", "订单量", "3132"]], "goals": "[1]问天气(User主动，User问天气，根据给定知识，Bot回复完整的天气信息，User满足并好评)-->[2]美食推荐(Bot主动推荐，这种天气温适合吃 『重庆辣子鸡』, User接受。需要聊2轮)-->[3]poi推荐(Bot主动，Bot推荐在 『老酒川菜坊宽院子店（原体南店）』 订 『重庆辣子鸡』, User问 『老酒川菜坊宽院子店（原体南店）』 的『人均价格』、『地址』、『评分』，Bot逐一回答后，最终User接受并提供预订信息:『就餐时间』 和 『就餐人数』)-->[4]再见", "user_profile": {"喜欢的电影": " 无人区", "职业状态": "学生", "喜欢的新闻": " 黄渤 的新闻", "拒绝": " 音乐", "同意的poi": " 老酒川菜坊宽院子店（原体南店）", "同意的美食": " 重庆辣子鸡", "接受的电影": [" 小王子", "每当变幻时", "寻龙诀", "人再囧途之泰囧", "疯狂的石头"], "性别": "女", "年龄区间": "小于18", "居住地": "石家庄", "喜欢的明星": " 黄渤", "没有接受的电影": " 上车，走吧", "姓名": "曾萍倩"}}
	\subsection{Data Collection}
	To obtain this data, we develop an interface and a pairing mechanism. We pair up task workers and give each of them a role of seeker or recommender. Then the two workers conduct data annotation with the help of task templates, seeker profiles and knowledge graph. In addition, we ask that the goals in templates must be tagged in every dialog. %Figure 3 provides a sample task template and an annotated dialog. Please see supplemental material for a screen snapshot of annotation interface.
	
	\textbf{Data structure} We organize the dataset of \emph{DuRecDial} according to seeker IDs. In \emph{DuRecDial}, there are multiple seekers (each with a different profile) and only one recommender. Each seeker $s_k$ has multiple dialogs $\{d^{s_k}_{i}\}_i$ with the recommender. For each dialog $d^{s_k}_{i}$, we provide a knowledge graph and a goal sequence for data annotation, and a seeker profile updated with this dialog. %In addition, we provide annotation information for goals in dialogs and the seeker's profile updated with ${d_i}^0_t$. 

	\begin{table}[t]
		\centering
		\small
		\begin{tabular}{ p{0.5in} p{1in}  r } 
			\toprule[1.0pt]
			\multirow{4}{0.5in}{Knowledge graph} & \#Domains &  7\\  
			&\#Entities &  21,837\\  
			&\#Attributes &  454\\  
			&\#Triples &  222,198\\ 
			\hline
			\multirow{4}{0.3in}{DuRecDial} &\#Dialogs &  10,190\\
			&\#Sub-dialogs for QA/Rec/task/chitchat & 6,722/8,756/3,234/10,190\\
			&\#Utterances &  155,477 \\
			&\#Seekers &  1362 \\
			&\#Entities recommended/accepted/rejected& 11,162/8,692/2,470/ \\ 
			\bottomrule[1.0pt]
			
		\end{tabular}
		\caption{Statistics of knowledge graph and DuRecDial.}
		\label{table:statistics}
	\end{table}

	\textbf{Data statistics} Table \ref{table:statistics} provides statistics of knowledge graph and \emph{DuRecDial}, indicating rich variability of dialog types and domains. 
	
	\textbf{Data quality} We conduct human evaluations for data quality. A dialog will be rated "1" if it follows the instruction in task templates and the utterances are fluent and grammatical, otherwise "0". Then we ask three persons to judge the quality of 200 randomly sampled dialogs. Finally we obtain an average score of 0.89 on this evaluation set. %Moreover, the ratio of dialogs with 1 or 0 is $90\%$ or $10\%$. 

	\begin{figure*}[t]
		%\centering\includegraphics[width=1.8in]{Figure2.eps}
		%\centering\includegraphics[width=3.1in]{Figure2.eps}
		%\centering\includegraphics[width=2.8in]{Figure2.pdf}
		%\centering\includegraphics[height=3.5in,width=6.5in]{./FIGURE/model_new.pdf}
		%\centering\includegraphics[height=3.5in,width=6.5in]{./FIGURE/acl2020_model.pdf}
		\centering\includegraphics[height=3.9in,width=6.6in]{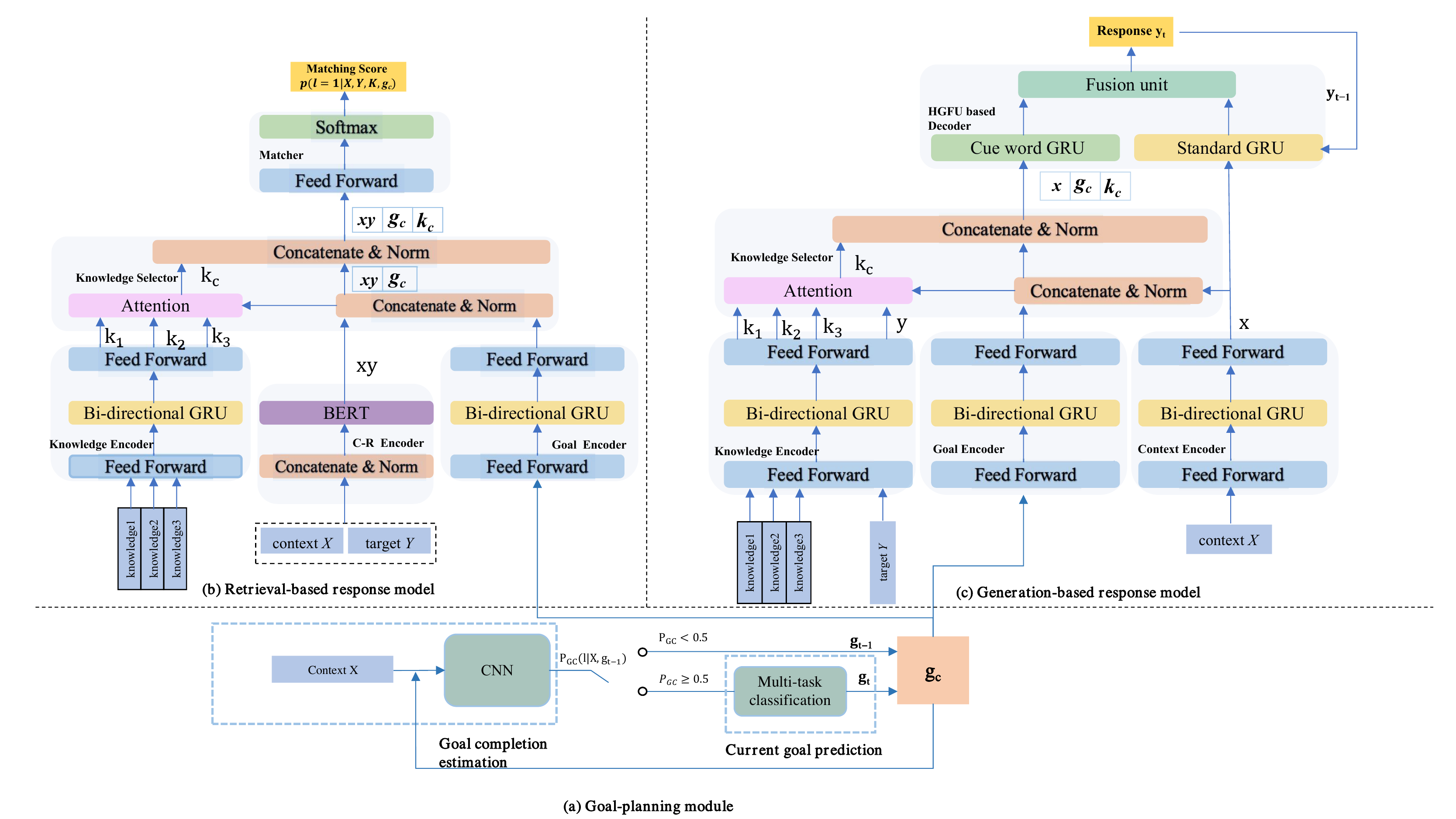}
		%\centering\includegraphics[width=6in]{acl2020/model_new.pdf}
		%\caption{\footnotesize{The architecture of AKGCM and RL based knowledge selector.RL based knowledge selector learns to traverse the graph from the starting vertex and finally stop at the answer vertex .}}\label{fig:architecture}
		\caption{The architecture of our multi-goal driven conversation generation framework (denoted as \textbf{MGCG}).}\label{fig:architecture}
	\end{figure*}

	\section{Our Approach}
	\label{sec:model}
	\subsection{Problem Definition and Framework Overview}
	\label{sec:3.1}
	\textbf{Problem definition} Let $D^{s_k}=\{d^{s_k}_{i}\}^{N_{D^{s_k}}}_{i=0}$ denote a set of dialogs by the seeker $s_k$ $(0\le k<N_s)$, where $N_{D^{s_k}}$ is the number of dialogs by the seeker $s_k$, and $N_s$ is the number of seekers. Recall that we attach each dialog (say $d^{s_k}_{i}$) with an updated seeker profile (denoted as $\mathcal{P}^{s_k}_{i}$), a knowledge graph $\mathcal{K}$, and a goal sequence $\mathcal{G}=\{g_{t}\}^{T_g-1}_{t=0}$. Given a context $X$ with utterances $\{u_j\}_{j=0}^{m-1}$ from the dialog $d^{s_k}_{i}$, a goal history $\mathcal{G'}=\{g_{0},...,g_{t-1}\}$ ($g_{t-1}$ as the goal for $u_{m-1}$), $\mathcal{P}^{s_k}_{i-1}$ and $\mathcal{K}$, the aim is to provide an appropriate goal $g_{c}$ to determine where the dialog goes and then produce a proper response $Y=\{y_{0}, y_{1}, ..., y_{n}\}$ for completion of the goal $g_{c}$.

	\textbf{Framework overview} The overview of our framework MGCG is shown in Figure \ref{fig:architecture}. The goal-planning module outputs goals to proactively and naturally lead the conversation. It first takes as input $X$, $\mathcal{G'}$, $\mathcal{K}$ and $\mathcal{P}^{s_k}_{i-1}$, then outputs $g_{c}$. The responding module is responsible for completion of each goal by producing responses conditioned on $X$, $g_{c}$, and $\mathcal{K}$. For implementation of the responding module, we adopt a retrieval model and a generation model proposed by \citet{Wu2019}, and modify them to suit our task. 
	
	For model training, each [context, response] in $d^{s_k}_{i}$ is paired with its ground-truth goal, $\mathcal{P}^{s_k}_{i}$ and $\mathcal{K}$. These goals will be used as answers for training of the goal-planning module, while the tuples of [context, a ground-truth goal, $\mathcal{K}$, response] will be used for training of the responding module.
	
	\subsection{Goal-planning Model}
	As shown in Figure ~\ref{fig:architecture}(a), we divide the task of goal planning into two sub-tasks, goal completion estimation, and current goal prediction. 
	
	\textbf{Goal completion estimation} For this subtask, we use Convolutional neural network (CNN)\cite{kim-2014} to estimate the probability of goal completion by:
	%For this subtask, we estimate the probability of goal completion by: 
	\begin{equation}
	\small
	P_{GC} (l=1|X, g_{t-1}). 
	\end{equation}
	
	\textbf{Current goal prediction} If $g_{t-1}$ is not completed ($P_{GC}< 0.5$), then $g_{c}=g_{t-1}$, where $g_{c}$ is the goal for $Y$. Otherwise, we use CNN based multi-task classification to predict the current goal by maximizing the following probability: 
	\begin{equation}
	\small
	g_{t} = \arg\max_{g^{ty}, g^{tp}} P_{GP} (g^{ty},g^{tp}|X, \mathcal{G'},\mathcal{P}^{s_k}_{i}, \mathcal{K}),
	\end{equation}
	\begin{equation}
	\small
	g_{c}=g_{t},
	\end{equation}
	where $g^{ty}$ is a candidate dialog type and $g^{tp}$ is a candidate dialog topic.
	
	\subsection{Retrieval-based Response Model}
	
	In this work, conversational goal is an important guidance signal for response ranking. Therefore, we modify the original retrieval model to suit our task by emphasizing the use of goals.%The original retrieval-based model in \cite{Wu2019} tries to select the best response from a set of candidates with the use of dialogue context and selected knowledge. 
	
	As shown in Figure ~\ref{fig:architecture}(b), our response ranker consists of five components: a context-response representation module (C-R Encoder), a knowledge representation module (Knowledge Encoder), a goal representation module (Goal Encoder), a knowledge selection module (Knowledge Selector), and a matching module (Matcher). 
	
	The C-R Encoder has the same architecture as BERT  \cite{devlin2018bert}, and it takes a context $X$ and a candidate response $Y$ as segment\_a and segment\_b in BERT, and leverages a stacked self-attention to produce the joint representation of $X$ and $Y$, denoted as $xy$. 
	
	Each related knowledge $knowledge_i$ is also encoded as a vector by the Knowledge Encoder using a bi-directional GRU \cite{chung2014empirical}, which can be formulated as $k_i = [\overrightarrow{h_{T_k}};\overleftarrow{h_0}]$, where $T_k$ denotes the length of knowledge, $\overrightarrow{h_{T_k}}$ and $\overleftarrow{h_0}$ represent the last and initial hidden states of the two directional GRU respectively. 
	
	The Goal Encoder uses bi-directional GRUs to encode a dialog type and a dialog topic for goal representation (denoted as $g_{c}$).
	
	For knowledge selection, we make the context-response representation $xy$ attended to all knowledge vectors ${k_i}$ and get the attention distribution: 
	\begin{equation}
	\small
	p(k_i|x,y,g_{c}) = \frac{exp( \textbf{MLP}([xy;g_{c}]) \cdot k_i )}{\sum_j exp( \textbf{MLP}([xy;g_{c}]) \cdot k_j )}
	\end{equation}
	We fuse all related knowledge information into a single vector $k_c = \sum_i p(k_i|x,y,g_{c}) * k_i$. 
	
	We view $k_c$, $g_c$ and $xy$ as the information from knowledge source, goal source and dialogue source respectively, and fuse the three information sources into a single vector via concatenation. Finally we calculate a matching probability for each $Y$ by:
	\begin{equation}
	\small
	p(l=1|X,Y,K,g_{c}) = softmax(\textbf{MLP}([xy;k_c;g_{c}]))
	\end{equation}
	\noindent
	%Thus this goal-driven response ranker can fully leverage the context, the conversational goal and the related knowledge for response ranking. 
	
	\subsection{Generation-based Response Model}
	
	To highlight the importance of conversational goals, we also modify the original generation model by introducing an independent encoder for goal representation. As shown in Figure ~\ref{fig:architecture}(c), our generator consists of five components: a Context Encoder, a Knowledge Encoder, a Goal Encoder, a Knowledge Selector, and a Decoder. 
	
	Given a context $X$, conversational goal $g_{c}$ and knowledge graph $K$, our generator first encodes them as vectors with the use of above encoders (based on bi-directional GRUs). %Especially, the dialogue context $X$ and dialogue goal $G$ are fused into the same vector $x$ by sequentially concatenate $G$ and $X$ into a single sentence, then feed to the encoder. 
	
	We assume that using the correct response will be conducive to knowledge selection. Then minimizing KLDivloss will make the effect of knowledge selection in the prediction stage (not use the correct response) close to that of knowledge selection with correct response. For knowledge selection, the model learns knowledge-selection strategy through minimizing the KLDivLoss between two distributions, a \emph{prior distribution} $p(k_i | x,g_{c})$ and a \emph{posterior distribution} $p(k_i | x,y,g_{c})$. It is formulated as:
	\begin{equation} 
	\small
	p(k_i | x, y, g_{c}) = \frac{exp(k_i \cdot \textbf{MLP}([x;y;g_{c}]))}{\sum_{j=1}^N exp(k_j \cdot \textbf{MLP}([x;y;g_{c}]))}
	\end{equation}
	\begin{equation} 
	\small
	p(k_i | x, g_{c}) = \frac{exp(k_i \cdot \textbf{MLP}([x;g_{c}])}{\sum_{j=1}^N exp(k_j \cdot \textbf{MLP}([x;g_{c}])}
	\end{equation}
	\begin{equation} 
	\small
	L_{KL}(\theta) = \frac{1}{N} \sum_{i=1}^N p(k_i | x,y,g_{c}) log \frac{p(k_i | x,y,g_{c})}{p(k_i | x,g_{c})}
	\end{equation}
	\noindent

	In training procedure, we fuse all related knowledge information into a vector $k_c=\sum_i p(k_i|x,y,g_{c})*k_i$, same as the retrieval-based method, and feed it to the decoder for response generation. In testing procedure, the fused knowledge is estimated by $k_c=\sum_i p(k_i|x,g_{c})*k_i$ without ground-truth responses. The decoder is implemented with the \emph{Hierarchical Gated Fusion Unit} described in \cite{yao2017towards}, which is a standard GRU based decoder enhanced with external knowledge gates. In addition to the loss $L_{KL}(\theta)$, the generator uses the following losses:
	\begin{description}
		\item \textbf{NLL Loss}: It computes the negative log-likelihood of the ground-truth response ($L_{NLL}(\theta)$). 
		\item \textbf{BOW Loss}: We use the BOW loss proposed by \citet{zhao2017learning}, to ensure the accuracy of the fused knowledge $k_c$ by enforcing the relevancy between the knowledge and the true response.\footnote{The BOW loss is to introduce an auxiliary loss that requires the decoder network to predict the bag-of-words in the response to tackle the vanishing latent variable problem.} Specifically, let $w = \textbf{MLP}(k_c) \in \mathcal{R}^{|V|}$, where $|V|$ is vocabulary size. We define:
		\begin{equation} 
			\small
			p(y_t|k_c) = \frac{exp(w_{y_t})}{\sum_{v=1}^V exp(w_v)}.
		\end{equation}
		%\begin{equation} 
		%\small
		%p(y_t | k_c)= \frac{exp(w_y_t)}{\sum_{v}^V exp(w_v)}.
		%\end{equation}
		
		\noindent
		Then, the BOW loss is defined to minimize:
		\begin{equation} 
			\small
			L_{BOW}(\theta) = - \frac{1}{m} \sum_{t=1}^m log p(y_t | k_c)
		\end{equation}
	\end{description}
	\noindent
	
	Finally, we minimize the following loss function:
	\begin{equation}
		\small
		L(\theta) = \alpha \cdot L_{KL}(\theta)  + \alpha \cdot L_{NLL}(\theta)  +  L_{BOW}(\theta) 
	\end{equation}
	where $\alpha$ is a trainable parameter.
	
	\begin{table*}
		\centering
		\small
		\begin{tabular}{c c c c c c }%{ r r r r r r r r r r  } 
			\toprule[1.0pt]
			%Methods$\downarrow$ Metrics$\rightarrow$  &  \emph{Hits{@}1}/\emph{Hits{@}3}  &F1/ BLEU2 & PPL & DIST-2 & Kg. P/R/F1 &$Acc_{gc}$ & $Acc_{ty}$/$Acc_{tp}$   \\  
			Methods$\downarrow$ Metrics$\rightarrow$  &  \emph{Hits{@}1}/\emph{Hits{@}3}  &F1/ BLEU2 & PPL & DIST-2 & Knowledge P/R/F1  \\  
			\toprule[1.0pt]
			%Goal planning & -/- & -                  &   - &  - & - &  92.61\%&93.82\%/41.03\%\\ \hline
			S2S- gl.- kg.     & 6.78\% / 24.55\% & 23.97 / 0.065             &   27.31 &  0.011 & 0.275 / 0.209 / 0.216  \\  
			S2S+gl.- kg.       & 8.03\% / 27.71\% & 24.78 / 0.077             &   24.82 &  0.012 & 0.287 / 0.223 / 0.231  \\   
			S2S+gl.+kg.       & 8.37\% / 27.67\% & 24.66 / 0.072             &   23.96&  0.011 & 0.295 / 0.239 / 0.253  \\ \hline
			MGCG\_R- gl.- kg. & 19.58\% / 42.75\% & 33.22 / 0.207             &   - &  0.171 &  0.344 / 0.301 / 0.306  \\  
			MGCG\_R+gl.- kg.  & 19.77\% / 42.99\% & 33.78 / 0.223             &   - &  0.185 &  0.351 / 0.322 / 0.309  \\
			MGCG\_R+gl.+kg.   & \textbf{20.33}\% / \textbf{43.61}\% & 33.93 / \textbf{0.232}             &   - &  \textbf{0.187} & 0.349 / 0.331 / 0.316  \\ \hline
			MGCG\_G- gl.- kg. & 13.26\% / 36.07\% & 33.11 / 0.189             &   18.51 &  0.037 & 0.386 / 0.349 / 0.358  \\ 
			MGCG\_G+gl.- kg.  & 14.21\% / 38.91\% & 35.21 / 0.213             &   17.78 &  0.049 & 0.393 / 0.352 / 0.351  \\ 
			MGCG\_G+gl.+kg.   & 14.38\% / 39.70\% & \textbf{36.81} / 0.219             &   \textbf{17.69} &  0.052 & \textbf{0.401} / \textbf{0.377} / \textbf{0.383}  \\ 
			\bottomrule[1.0pt]
			
		\end{tabular}
		\caption{Automatic evaluation results. +(-)gl. represents ``with(without) conversational goals''. +(-)kg. represents ``with(without) knowledge''. For ``S2S +gl.+kg.'', we simply concatenate the goal predicted by our model, all the related knowledge and the dialog context as its input.}
		\label{table:auto-results}
	\end{table*}
	
	\begin{table*}
		\centering
		\small
		\begin{tabular}{  r r r r r  r r } 
			\toprule[1.0pt]
			& \multicolumn{4}{c|}{Turn-level results}  & \multicolumn{2}{c}{Dialog-level results} \\ \cmidrule{2-5} \cmidrule{6-7}
			Methods$\downarrow$ Metrics$\rightarrow$ & Fluency  & Appro. & Infor. & \multicolumn{1}{c|}{Proactivity}  &  Goal success rate & Coherence \\ 
			%Scores$\rightarrow$ & (0,1,2) & (0,1,2) & (0,1,2) &  (0,1,2) & (0,1,2)  & (0,1,2) &  (0,1,2)  \\ \hline
			\toprule[1.0pt] 
			%Goal planning model& - & - & - & - & & -  \\ \hline
			%S2S - gl. - kg.    & 00 & 00 & 00 & 00 & 00 & 00 & - \\ 
			S2S +gl. +kg.      & 1.08 & 0.23 & 0.37 & 0.94 & 0.37 & 0.49   \\ \hline
			%MGCG\_R - gl. - kg.& 00 & 00 & 00 & 00 & 00 & 00 & - \\ 
			MGCG\_R +gl. +kg.  & \textbf{1.98} & 0.60 & 1.28 & 1.22 & 0.68 & 0.83  \\ \hline
			%MGCG\_G - gl. - kg.& 00 & 00 & 00 & 00 & 00 & 00   \\ 
			MGCG\_G +gl. +kg.  & 1.94 & \textbf{0.75} & \textbf{1.68} & \textbf{1.34} & \textbf{0.82} & \textbf{0.91}   \\ 
			\bottomrule[1.0pt]
		\end{tabular}
		\caption{Human evaluation results at the level of turns and dialogs.}
		\label{table:human-results}
	\end{table*}

	\section{Experiments and Results}
	
	\subsection{Experiment Setting}
	We split \emph{DuRecDial} into train/dev/test data by randomly sampling 65\%/10\%/25\% data at the level of seekers, instead of individual dialogs. To evaluate the contribution of goals, we conduct an ablation study by replacing input goals with “UNK” for responding model. For knowledge usage, we conduct another ablation study, where we remove input knowledge by replacing them with “UNK”. 
	
	\subsection{Methods\footnote{Please see Appendix 2. for model parameter settings.}}

	\textbf{S2S} We implement a vanilla sequence-to-sequence model \cite{Sutskever2014}, which is widely used for open-domain conversation generation. 
	
	\textbf{MGCG\_R}: Our system with automatic goal planning and a retrieval based responding model. 
	
	\textbf{MGCG\_G}: Our system with automatic goal planning and a generation based responding model.

	\subsection{Automatic Evaluations}
	\textbf{Metrics} For automatic evaluation, we use several common metrics such as BLEU \cite{Papineni2002}, F1, perplexity (PPL), and DISTINCT (DIST-2) \cite{li2016diversity} to measure the relevance, fluency, and diversity of generated responses. Following the setting in previous work \cite{Wu2019,zhang2018personalizing}, we also measure the performance of all models using Hits@1 and Hits@3.\footnote{Candidates (including golden response) are scored by PPL using the generation-based model, then candidates are sorted based on the scores, and Hits@1 and Hits@3 are calculated.} Here we let each model to select the best response from 10 candidates. Those 10 candidate responses consist of the ground-truth response generated by humans and nine randomly sampled ones from the training set. Moreover, we also evaluate the knowledge-selection capability of each model by calculating knowledge precision/recall/F1 scores as done in \citet{Wu2019}.\footnote{When calculating the knowledge precision/recall/F1, we compare the generated results with the correct knowledge.} In addition, we also report the performance of our goal planning module, including the accuracy of goal completion estimation, dialog type prediction, and dialog topic prediction. 
	
	\textbf{Results} Our goal planning model can achieve accuracy scores of 94.13\%, 91.22\%, and 42.31\% for goal completion estimation, dialog type prediction, and dialog topic prediction. The accuracy of dialog topic prediction is relatively low since the number of topic candidates is very large (around 1000), leading to the difficulty of topic prediction. As shown in Table \ref{table:auto-results}, for response generation, both MGCG\_R and MGCG\_G outperform S2S by a large margin in terms of all the metrics under the same model setting (without gl.+kg., with gl., or with gl.+kg.). Moreover, MGCG\_R performs better in terms of Hits{@}k and DIST-2, but worse in terms of knowledge F1 when compared to MGCG\_G.\footnote{We calculate an average of F1 over all the dialogs. It might result in that the value of F1 is not between P and R.} It might be explained by that they are optimized on different metrics. We also found that the methods using goals and knowledge outperform those without goals and knowledge, confirming the benefits of goals and knowledge as guidance information. %Specifically, background knowledge has a greater impact on the MGCG\_G model, confirming the effectiveness of posterior knowledge selection strategy.
	
	\subsection{Human Evaluations}
	\textbf{Metrics:} The human evaluation is conducted at the level of both turns and dialogs. 
	
	For turn-level human evaluation, we ask each model to produce a response conditioned on a given context, the predicted goal and related knowledge.\footnote{Please see Appendix 3. for more details.} The generated responses are evaluated by three annotators in terms of fluency, appropriateness, informativeness, and proactivity. The appropriateness measures if the response can complete current goal and it is also relevant to the context. The informativeness measures if the model makes full use of knowledge in the response. The proactivity measures if the model can successfully introduce new topics with good fluency and coherence.
	
	For dialogue-level human evaluation, we let each model converse with a human and proactively make recommendations when given the predicted goals and related knowledge.\footnote{Please see Appendix 4. for more details.} For each model, we collect 100 dialogs. These dialogs are then evaluated by three persons in terms of two metrics: (1) goal success rate that measures how well the conversation goal is achieved, and (2) coherence that measures relevance and fluency of a dialog as a whole.
	
	All the metrics has three grades: good(2), fair(1), bad(0). For proactivity, ``2'' indicates that the model introduces new topics relevant to the context, ``1'' means that no new topics are introduced, but knowledge is used, ``0'' means that the model introduces new but irrelevant topics. For goal success rate, ``2'' means that the system can complete more than half of the goals from goal planning module, ``0'' means the system can complete no more than one goal, otherwise ``1''. For coherence, ``2''/``1''/``0'' means that two-thirds/one-third/very few utterance pairs are coherent and fluent. %Please see supplemental material for more details about all metrics. %For goal-seq quality metric, ``2''/``1''/``0'' means that there are no/only one/more than one goals being inappropriate for given dialog context. Please see supplemental material for more details about other metrics.
	
	\textbf{Results} All human evaluations are conducted by three persons. As shown in Table \ref{table:human-results}, our two systems outperform S2S by a large margin, especially in terms of appropriateness, informativeness, goal success rate and coherence. In particular, S2S tends to generate safe and uninformative responses, failing to complete goals in most of dialogs. Our two systems can produce more appropriate and informative responses to achieve higher goal success rate with the full use of goal information and knowledge. Moreover, the retrieval-based model performs better in terms of fluency since its response is selected from the original human utterances, not automatically generated. But it performs worse on all the other metrics when compared to the generation-based model. It might be caused by the limited number of retrieval candidates. Finally, it can be seen that there is still much room for performance improvement in terms of appropriateness and goal success rate, which will be left as the future work.
	
	\begin{table}[t]
		\centering
		\small
		\begin{tabular}{ r r r r r r } 
			\toprule[1.0pt]
			\multicolumn{2}{r}{Methods$\rightarrow$}   & S2S   & MGCG\_R   & MGCG\_G \\ 
			\multicolumn{2}{r}{Metrics$\downarrow$Types$\downarrow$}   &  +gl. +kg.   &  +gl. +kg. &  +gl. +kg. \\ \toprule[1.0pt] 
			\multirow{5}{0.5cm}{\#Failed gl./ \#Completed gl.}  & Rec. &106/7 &95/18 &93/\textbf{20} \\  
			& Chitchat & 120/93 &96/117 &80/\textbf{133} \\  
			& QA & 66/5 &61/10 &60/\textbf{11} \\  
			& Task &45/4 &36/\textbf{13} &39/10 \\  
			& Overall &337/109 &288/158 &272/\textbf{174} \\ \hline
			\multirow{5}{0.5cm}{\#Used kg.}  & Rec. &0 &\textbf{8} &7 \\   
			& Chitchat &9 &25 &\textbf{33} \\  
			& QA &5 &10 &\textbf{15} \\  
			& Task &0 &\textbf{3} &2 \\
			& Overall &14 &46 &\textbf{57} \\ 
			\bottomrule[1.0pt]
		\end{tabular}
		\caption{Analysis of goal completion and knowledge usage across different dialog types.}
		\label{table:analysis-results}
	\end{table}
	\subsection{Result Analysis} 
	In order to further analyze the relationship between knowledge usage and goal completion, we provide the number of failed goals, completed goals, and used knowledge for each method over different dialog types in Table ~\ref{table:analysis-results}. We see that the number of used knowledge is proportional to goal success rate across different dialog types or different methods, indicating that the knowledge selection capability is crucial to goal completion through dialogs. Moreover, the goal of chitchat dialog is easier to complete in comparison with others, and QA and recommendation dialogs are more challenging to complete. How to strengthen knowledge selection capability in the context of multi-type dialogs, especially for QA and recommendation, is very important, which will be left as the future work.

	\section{Conclusion}
	We identify the task of conversational recommendation over multi-type dialogs, and create a dataset \emph{DuRecDial} with multiple dialog types and multi-domain use cases. We demonstrate usability of this dataset and provide results of state of the art models for future studies. The complexity in \emph{DuRecDial} makes it a great testbed for more tasks such as knowledge grounded conversation \cite{Ghazvininejad2018}, domain transfer for dialog modeling, target-guided conversation \cite{tang2019target} and multi-type dialog modeling \cite{Yu2017}. The study of these tasks will be left as the future work.
	%, and 
	%	\clearpage
	
\section*{Acknowledgments}

We would like to thank Ying Chen for dataset annotation and thank Yuqing Guo and the reviewers for their insightful comments. This work was supported by the National Key Research and Development Project of China (No. 2018AAA0101900) and the Natural Science Foundation of China (No. 61976072).

\bibliography{anthology,acl2020}
\bibliographystyle{acl_natbib}

\appendix

%\clearpage
%\appendixpage
\section*{Appendix}
%\appendix

\subsection*{1. Dataset collection process}

\subsection*{1.1 Collection of seeker profiles/knowledge graph/task templates}

\paragraph{Collection of seeker profile}
The attributes of seeker profiles are shown as follows: name, gender, age range, city of residence, occupation status, and seeker preference. Seeker preference includes : domain preference, seed entity preference, entity list rejected by the seeker,  entity list accepted by the seeker.

\begin{itemize}
\item\textbf{Name}: We generate the first-name Chinese character(or last-name Chinese character) by randomly sampling Chinese characters from a set of candidate characters used as first name (or last name) for the gender of the seeker.

\item\textbf{Gender}: We randomely select "male" or "female" as the seeker's gender.

\item\textbf{Age range}: We randomly choose one from the 5 age ranges.

\item\textbf{Residential city}: We randomly choose one from 55 cities in China as the seeker's residential city.

\item\textbf{Occupation status}: We randomly choose one from "student", "worker" and "retirement" based on above age range.

\item\textbf{Domain preference}: We randomly select one or two domains as ones that the seeker likes (e.g., movie, food), and one domain as the one that the seeker dislikes (e.g., news). It will affect the setting of task templates for this seeker.

\item\textbf{Seed entity preference}: We randomly select one or two entities from KG entities of the domains preferred by the seeker as his/her preference at entity level. It will affect the setting of task templates for this seeker.

\item\textbf{Rejected entity list and accepted entity list}: Both of them are empty at the beginning, and they will be updated as the conversation progresses; the two lists will affect the recommendation results of subsequent conversations to some extent.
\end{itemize}

\paragraph{Collection of Knowledge graph (KG)}
The domain of knowledge graph include stars, movies, music, news, food, POI(Point of Interest), weather.

\begin{itemize}
	\item\textbf{Stars}: including the introduction, achievements, awards, comments, birthday, birthplace, height, weight, blood type, constellation, zodiac, nationality, friends, etc
	
	\item\textbf{Film}: including film rating, comments, region, leading role, director, category, evaluation, award, etc
	
	\item\textbf{Music}: singer information, comments, etc
	
	\item\textbf{News}: including the topic, content, etc
	
	\item\textbf{Food}: including the name, ingredients, category, etc
	
	\item\textbf{POI}: including restaurant name, average price, score, order quantity, address, city, specialty, etc
	
	\item\textbf{Weather}: historical weather of 55 cities from July 2017 to August 2019
\end{itemize}

\paragraph{Collection of task templates} First we manually annotate a list of around 20 high-level goal sequences as candidates. Most of these goal sequences include 3 to 5 high-level goals. Here each high-level goal contains a dialog type and a domain (not entity or chatting topic). Then for each seeker, we select appropriate high-level goal sequences from the above list, which contains the domains that fall into the seeker's preferred domain list. 

To collect goal sequences at entity level, we first use the seed entities of the seeker to enrich the information of above high-level goal sequences. If the seed entities are not enough, or there is no seeds for some domains in the high-level goal sequences, we select some entities from KG for each goal domain based on embedding based similarity scores of the seed entities (of current seeker) and the candidate entity. Then we obtain goal sequences at entity level. Finally we use some rules to generate a description for each goal (e.g., which side, the seeker or the recommender, to start the dialog, how to complete the goal). Thus we have task templates for guidance of data annotation.

To introduce diverse interaction behavior for recommendation, we design some fine-grained interaction operations, e.g., the seeker may reject the initial recommendation, or mention a new topic, or ask a question about an entity, or simply accept the recommendation. Each interaction operation corresponds to a goal. We randomly sample one of above operations and insert it into the entity-level goal sequences to diversify recommendation dialogs. The entities associated with the above interaction operations are selected from the KG based on their similarity scores with current seeker's seed entites. If the entity will be accepted by the seeker as described in the task templates (including entity-level goal sequence and its description), then its similarity score with the seeker's seed entites should be relatively high. If the entity will be rejected by the seeker as described in the task templates, then its similarity score with the seeker's seed entites should be relatively low.

\subsection* {1.2 Dataset annotation process} We first release a small amount of data for training, and then carry out video training for annotation problems. After that,  a small amount of data is released again to select the final task workers. To ensure that at least two workers enter the task at the same time, we arrange multiple workers to log in the annotation platform. During annotation, each conversation is randomly assigned to two workers, one of whom plays the role of BOT and the other plays the role of User. Two workers conduct annotation based on the seeker profile, knowledge graph and task templates.

Due to the complexity of our task design, the quality of data annotation may have a high variance. To address this problem, we provide a strict data annotation standard to guide the workers to annotate in the way we expect them to be. After data annotation, multiple data specialists will review it. As long as one specialist thinks it does not meet the requirements, it will be marked back and re-annotation until all specialists think that it fully meets requirements.

\subsection* {2. Model Parameter Settings}
All models are implemented using PaddlePaddle.\footnote{It is an open source deep learning platform(https://www.paddlepaddle.org.cn/)} The parameters of all the modules are shown in Table ~\ref{table:analysis-1}.\footnote{Due to S2S model uses the same parameters as OpenNMT(https://github.com/OpenNMT/OpenNMT-py), its parameters are not listed.}

\subsection* {3. Turn-level Human Evaluation Guideline}

\textbf{Fluency} measures if the produced response itself is fluent: 
\begin{itemize}
\item score 0 (bad): unfluent and difficult to understand.
\item score 1 (fair): there are some errors in the response text but still can be understood.
\item score 2 (good): fluent and easy to understand. 
\end{itemize}

\noindent
\textbf{Appropriatenss} measures if the response can respond to the context: 
\begin{itemize}
\item score 0 (bad): Sub-dialogs for Rec and chitchat: not semantically relevant to the context or logically contradictory to the context.
Sub-dialogs for task-oriented: No necessary slot value is involved in the conversation.
Sub-dialogs for QA: Incorrect answer.
\item score 1 (fair): relevant to the context as a whole, but using some irrelevant knowledge, or not answering questions asked by the users.
\item score 2 (good): otherwise.
\end{itemize}

\begin{table}
\centering
\small
\begin{tabular}{ p{1.2in} r r} 
\toprule[1.0pt]
 {module} & Parameter & value\\ 
 \hline
Goal-planning model & Embedding Size & 256\\
& Hidden Size & 256\\
& Batch Size & 128\\
& Learning Rate & 0.002 \\
& Optimizer & Adam \\\hline
Retrieval-based model & Dropout& 0.1\\
& Embedding Size& 512\\
& Hidden Size & 512\\
& Batch Size& 32\\
& Learning Rate & 0.001\\
& Optimizer & Adam\\
& Weight Decay& 0.01\\
& Proportion Warmup& 0.1\\\hline
Generation-based model &Embedding Size &300\\
& Hidden Size & 800\\
& Batch Size & 16\\
& Learning Rate & 0.0005\\
& Grad Clip & 5\\
& Dropout & 0.2\\
& Beam Size & 10\\
& Optimizer & Adam\\
\hline
\end{tabular}
\caption{Model parameter settings.}
\label{table:analysis-1}
\end{table}

\noindent
\textbf{Informativeness} measures if the model makes full use of knowledge in the response:
 \begin{itemize}
 \item score 0 (bad): no knowledge is mentioned at all.
 \item score 1 (fair): only one knowledge triple is mentioned in the response.
 \item score 2 (good): more than one knowledge triple is mentioned in the response.\\
\end{itemize}

\noindent
\textbf{Proactivity} measures if the model can introduce new knowledge/topics in conversation:
\begin{itemize}
\item score -1 (bad): some new topics are introduced but irrelevant to the context.
\item score 0 (fair): no new topics/knowledge are used.
\item score 1(good): some new topics relevant to the context are introduced.\\
\end{itemize}

\noindent
\subsection* {4. Dialogue-level Human Evaluation Guideline}
\noindent
\textbf{Goal Completion} measures how good the given conversation goal is finished:
\begin{itemize}
\item score 0 (bad): less than half goals are achieved..
\item score 1 (fair): less than half goals are achieved with minor use of knowledge or goal information.
\item score 2 (good): more than half goals are achieved with full use of knowledge and goal information.\\
\end{itemize}

\noindent
\textbf{Coherence} measures the overall fluency of the whole dialogue:
\begin{itemize}
\item score 0 (bad): two-thirds responses irrelevant or logically contradictory to the previous context.
\item score 1 (fair): less than one-third responses irrelevant or logically contradictory to the previous context.
\item score 2 (good): very few response irrelevant or logically contradictory to the previous context.
\end{itemize}

\subsection* {5. Case Study}
\begin{figure*}[h]
\centering
\includegraphics[height=5in,width=6.5in]{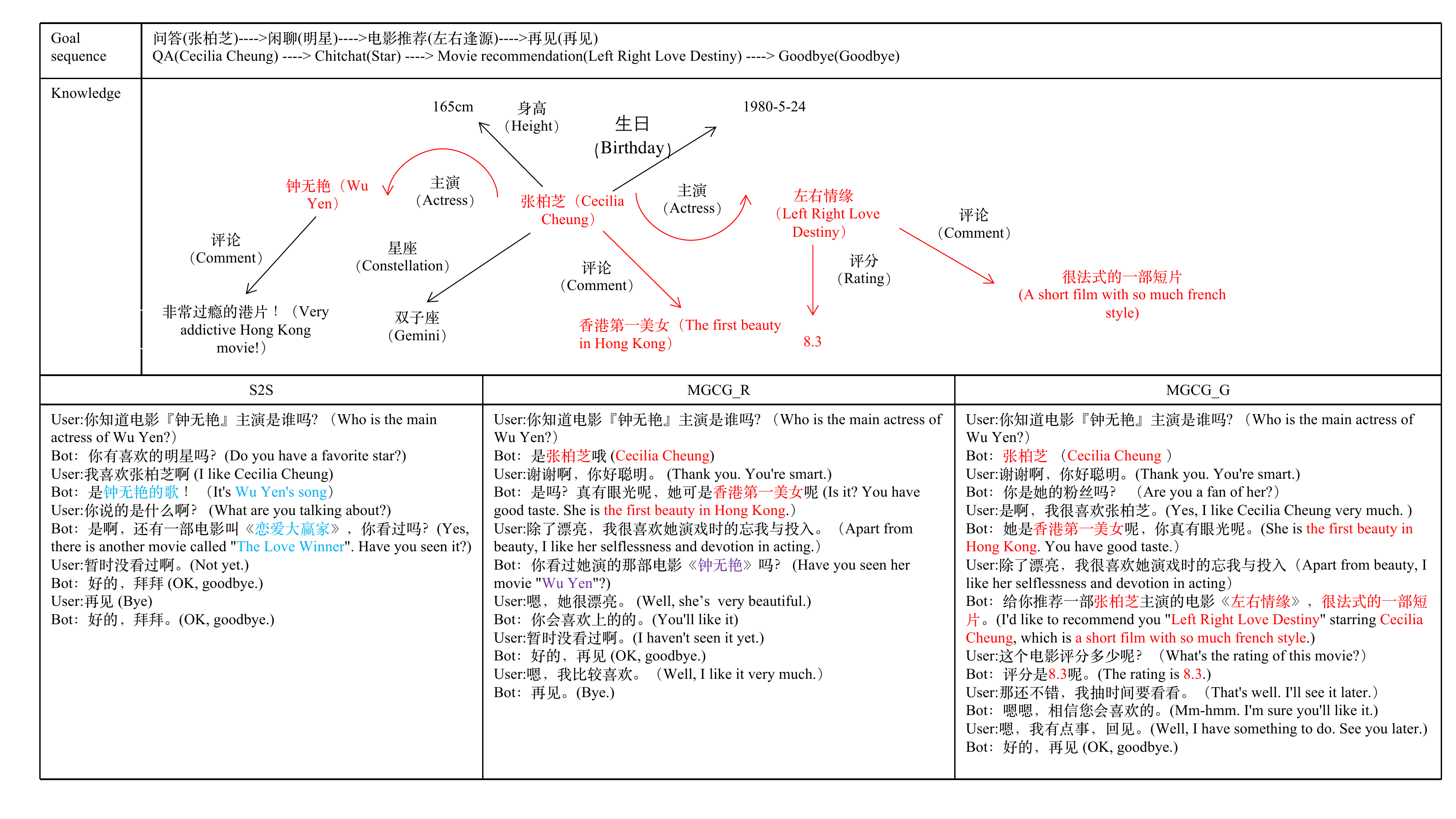}
\caption{Conversations generated by three different models: texts in red color represent correct knowledge being appropriate in current context, while texts in blue color represent inappropriate knowledge. Texts in purple color indicate that the use of knowledge is correct, but the response is not appropriate.}
\label{fig:case}
\end{figure*}

Figure ~\ref{fig:case} shows the conversations generated by the models via conversing with humans, given the conversation goal and the related knowledge.  It can be seen that our knowledge-aware generator can use more correct knowledge for diverse conversation generation. Even though the retrieval-based method can also produce knowledge-grounded responses, the used knowledge is relatively few and inappropriate. The seq2seq model can't successfully complete the given goal, as the knowledge is not fully used as our proposed knowledge-aware generator, making the generated conversation less diverse and sometimes dull.

\end{document}